\documentclass[11pt]{article}
\usepackage[preprint]{acl}
\usepackage{times}
\usepackage{latexsym}
\usepackage[T1]{fontenc}
\usepackage[utf8]{inputenc}
\usepackage{microtype}
\usepackage{amsmath,amssymb}
\usepackage{booktabs,array,multirow}
\usepackage[table]{xcolor}
\usepackage{colortbl}
\usepackage{soul}
\usepackage{algorithm}
\usepackage{algpseudocode}
\usepackage{graphicx}
\usepackage{listings}
\usepackage{tikz}
\usetikzlibrary{arrows.meta,positioning}
\newcommand{\method}{\textsc{EvoGuard}}
\definecolor{oursrow}{RGB}{229,238,251}
\definecolor{oursall}{RGB}{205,221,246}
\definecolor{allcol}{RGB}{253,245,230}
\definecolor{notebg}{RGB}{255,236,153}
\sethlcolor{notebg}

\makeatletter
\g@addto@macro\normalsize{%
  \setlength\abovedisplayskip{4pt plus 1pt minus 2pt}%
  \setlength\belowdisplayskip{4pt plus 1pt minus 2pt}%
  \setlength\abovedisplayshortskip{0pt plus 1pt}%
  \setlength\belowdisplayshortskip{3pt plus 1pt minus 1pt}}
\makeatother
\title{\method{}: Mitigating Degradation in Online Skill Evolution}
\makeatletter
\newcommand{\authormark}[1]{\textsuperscript{\ensuremath{#1}}}
\newcommand{\authornote}[2]{%
  \g@addto@macro\@thanks{%
    \begingroup
      \def\@thefnmark{\ensuremath{#1}}%
      \begin{NoHyper}\@footnotetext{#2}\end{NoHyper}%
    \endgroup}}
\makeatother
\author{%
  \textbf{Yanli Wang\authormark{1,2,3,\diamond},
  Zichen Bai\authormark{1,\diamond},
  Tianyue Jiang\authormark{2},
  Kunming Shao\authormark{4},
  Junhao Shi\authormark{1}}\\
  \textbf{Xueqin Liu\authormark{1},
  Jinguo Sun\authormark{1},
  Muye Li\authormark{1},
  Tingting Bi\authormark{3},
  Yanlin Wang\authormark{2}}\\
  \textbf{Shaoqiang Zheng\authormark{1,\dagger},
  Xuefei Diao\authormark{1},
  Zhuwei Zeng\authormark{1},
  Nan Duan\authormark{1},
  Peng Cao\authormark{1}}\\[0.35em]
  \authormark{1}JD.Com \quad
  \authormark{2}Sun Yat-sen University \quad
  \authormark{3}The University of Melbourne\\
  \authormark{4}The Hong Kong University of Science and Technology%
  \authornote{\diamond}{JD Tech Genius Team}%
  \authornote{\dagger}{Corresponding author}%
}
\begin{document}
\maketitle
\begin{abstract}
Skills let language agents reuse experience across tasks as natural-language documents. Recent work evolves skills from interaction, but existing methods have limitations when skills evolve online. Prior methods update a skill offline from a batch of trajectories collected over many tasks. Some then adopt the new skill without any check, which can degrade the skill over time. Others keep it only if it improves performance on a held-out validation set, which is too slow and costly to run after every update. We introduce \method{}, a pairwise skill selector that mitigates skill degradation at low latency and cost. We first run the agent iteratively to build pools of skills and score each skill on a private test set to obtain utility labels. We pair the skills and use rejection sampling to keep only pairs with a clear utility gap. The selector is trained on the logits of two label tokens to pick the better skill. At deployment it decides in a single forward pass, which is significantly faster than evaluation on the private test set. Compared with adopting every new skill, \method{} improves success by 11.0\% on ALFWorld, 5.9\% on ScienceWorld, and 2.7\% on LiveCodeBench.

\end{abstract}
\section{Introduction}
Language agents increasingly face streams of related tasks. Solving each task from scratch wastes what earlier interactions have revealed. A \emph{skill} packages such experience as a reusable natural-language document. It records procedures, decision rules, and recovery strategies. A frozen agent can then adapt through external text instead of weight updates \citep{wang2023voyager,anthropic2025agentskills,xu2026agentskillssurvey}. Recent work builds and evolves skills from interaction \citep{xia2026skillrl,yang2026skillopt,yao2026skillrise}. However, existing methods face three limitations when skills evolve online. In this setting a deployed agent updates its skill after every task.

\textbf{Gap 1: Offline multi-trajectory evolution does not match deployment.} Most methods update a skill offline from a batch of trajectories \citep{xia2026skillrl,yang2026skillopt}. The update runs in a training phase before deployment. It analyzes trajectories from many tasks at once. A deployed agent rarely works this way. Each task arrives once and is attempted once. The next task is usually a different one. The skill must therefore be revised from a single trajectory and then reused by later tasks. SkillRise carries a skill across tasks in this way \citep{yao2026skillrise}. However, it trains the executor jointly with skill curation and does not check individual updates.

\begin{figure*}[t]
\centering
\includegraphics[width=0.78\textwidth]{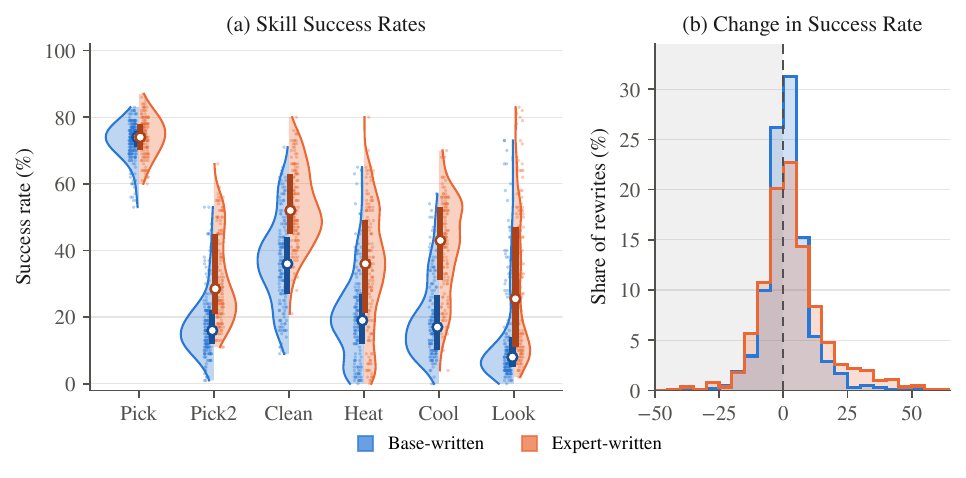}
\caption{(a) Success rate of the base-written and expert-written skills on the private test set for each task family. (b) Change in success rate after a single-trajectory rewrite (child minus parent). The shaded region marks degradation.}
\label{fig:pool}
\end{figure*}

\textbf{Gap 2: Adopting every new skill causes degradation.} A rewrite based on one trajectory can overfit that trajectory. A failed search may add a rule tied to one room. Even a successful episode may add redundant or misleading advice. This risk is measurable. We collect single-trajectory rewrites on ALFWorld and score every skill on a private test set of its task family (Figure~\ref{fig:pool}).

Figure~\ref{fig:pool}(a) shows that the success rates of these skills spread over a wide range. Skills of very different quality therefore coexist. Figure~\ref{fig:pool}(b) shows the change in success rate after one rewrite, computed as the rewrite minus the skill it came from. The empty skill also counts as a parent. Among the 1{,}137 rewrites written by the base generator, 42.2\% score below their parent. Another 7.7\% tie, and 50.0\% improve. A rewrite is thus nearly neutral on average. However, 7.7\% of rewrites lose at least 10\%, and 11.3\% gain at least 10\%. A stronger generator does not remove the problem. Of the 900 rewrites written by an expert generator, which is trained with reinforcement learning, 40.0\% also score below their parent. Adopting every rewrite therefore lets the skill drift. This motivates an explicit acceptance step.

\textbf{Gap 3: Validating every update is too costly.} Some methods guard against bad updates \citep{alzubi2026evoskill,yang2026skillopt}. They keep a new skill only if it improves performance on a held-out validation set. This requires many extra rollouts for every update. Each check thus costs far more than the task that triggered the update. An online agent cannot afford this cost after every task.

To address these gaps, we propose \method{}, a pairwise skill selector that amortizes validation. Tasks are grouped into task families, and each family keeps one skill. After each task, a fixed generator writes a new skill from the latest trajectory. \method{} then decides whether this new skill replaces the current skill. This decision only requires the relative order of the two skills, not their absolute utilities. We therefore cast it as a pairwise preference problem. We pay the cost of validation once before deployment and use it to train the selector. After that, each decision only compares two skill texts.

\method{} works in three stages (Figure~\ref{fig:overview}). \textbf{(1) Multi-source skill pool construction.} We run the rewrite loop on training tasks with a base generator and an expert generator, and collect the generated skills into a pool. \textbf{(2) Utility-grounded pairwise preference learning.} We estimate the utility of every skill on a private test set of its family. Rejection sampling keeps pairs from the same family with a clear utility gap. We then train a pairwise selector to predict which skill in a pair has the higher utility. \textbf{(3) Online skill selection.} After every task, the frozen selector reads the family description and the two skills. It keeps the skill it predicts to be better. This takes one forward pass and runs no extra rollouts in the environment.

We evaluate \method{} on ALFWorld \citep{shridhar2021alfworld} with three executors. The baselines include running without skills and adopting every new skill. They also include random selection from the skill history and three LLM-Judge variants. On skill pairs unseen during training, \method{} picks the better skill 95.0\% of the time. Online, it outperforms every other method in all six settings. Its average success with the Qwen3-4B and Qwen3-8B executors is 56.5\%. Adopting every new skill reaches 45.5\%, and running without skills reaches 36.7\%. The Qwen3-32B executor contributes no training labels. With this executor \method{} still reaches 69.2\% against 54.9\% without skills. On ScienceWorld \citep{wang2022scienceworld}, \method{} raises success from 10.2\% to 15.3\%. Our main contributions are as follows.
\begin{itemize}
\setlength{\itemsep}{0pt}
\item We formulate online skill evolution as a per-update acceptance problem. After each single-trajectory rewrite, the agent must decide whether the new skill replaces the current skill. We show that this decision matters. On ALFWorld, 42\% of rewrites lower performance on a private test set.
\item We propose \method{}, a pairwise skill selector that amortizes validation. Validation runs once offline to supervise the selector. Each online decision then needs one forward pass over two skill texts. It generates no tokens and runs no extra rollouts.
\item \method{} achieves the highest success among all methods on ALFWorld, ScienceWorld, and LiveCodeBench. It also transfers to an executor unseen during selector training.
\end{itemize}
\section{Related Work}
\label{sec:related}
\paragraph{Skill evolution.}
Skills package reusable experience as executable code or structured documents \citep{wang2023voyager,anthropic2025agentskills,xu2026agentskillssurvey}. Recent work evolves skills from interaction. SkillRL integrates recursive skill evolution with reinforcement learning \citep{xia2026skillrl}. SkillOpt treats a skill as the external state of a frozen agent and optimizes it under feedback \citep{yang2026skillopt}. SkillRise is the closest work to ours \citep{yao2026skillrise}. It also carries an evolving skill across distinct tasks and trains the executor jointly with skill curation through reinforcement learning. However, our method keeps both the executor and the generator frozen. It only decides whether a single-trajectory rewrite should replace the current skill. This decision is learned from supervised pairwise labels without policy optimization. \method{} is thus a separate module that can be attached to any rewrite loop.

\paragraph{Guarding skill updates.}
Several methods check an update before they keep it. EvoSkill and SkillOpt keep an update only if it improves performance on a held-out validation set \citep{alzubi2026evoskill,yang2026skillopt}. SkillEvo adds a governance layer that repairs degraded skills instead of only rejecting them \citep{yan2026skillevo}. SkillEvoReg regularizes skill evolution against overfitting and tests each candidate for regressions \citep{nie2026skillevoreg}. These methods execute additional tasks for every update. However, our method pays the validation cost only once before deployment. Each online decision then needs only a single forward pass over two skill texts. Other work chooses which skill to read from a library during an episode \citep{li2026skillgate}. Our method instead decides whether a new skill should replace the current one. Appendix~\ref{sec:morerelated} discusses agent memory, learned skill writers, and further related work.
\section{Problem Setting}
\label{sec:setting}
Tasks arrive one at a time. Let $\mathcal{F}$ be the set of task families. The $t$-th task is $x_t$, and its family is $f_t\in\mathcal{F}$. Each family $f$ keeps one textual skill $s_f$. Every skill starts as the empty document $\varnothing$. A frozen executor $E$ attempts $x_t$ with the current skill $s_{f_t}$. It returns a trajectory $\tau_t$ and an outcome $r_t$ that measures task success:
\begin{equation}
(\tau_t,r_t)=E(x_t,s_{f_t}).
\end{equation}
A fixed generator $G$ then writes a new skill from the current skill, the trajectory, and the outcome:
\begin{equation}
\tilde{s}_t=G(s_{f_t},\tau_t,r_t).
\end{equation}
A format check $v(\tilde{s}_t)\in\{0,1\}$ tests whether the new skill is well formed. If $v(\tilde{s}_t)=1$, a selector $C$ chooses the next skill of the family:
\begin{equation}
s_{f_t}\leftarrow C(f_t,s_{f_t},\tilde{s}_t)\in\{s_{f_t},\tilde{s}_t\}.
\label{eq:select}
\end{equation}
Otherwise the current skill is kept. The skills of other families do not change. The selected skill serves later tasks of the same family. It cannot change the outcome of the task that has just finished.

\paragraph{Information boundary.}
The executor observes the current task and the environment feedback. The generator observes the current skill, the latest trajectory, and its outcome. The selector observes only the family name, a generic family description, and the two skill texts. It does not see the latest task, its trajectory, or its outcome.

\paragraph{Objective.}
Let $\mathcal{D}_f$ be the task distribution of family $f$. Let $r(x,s)$ denote the outcome of $E(x,s)$. We define the utility of a skill $s$ as the expected outcome of the executor with this skill:
\begin{equation}
u_f(s)=\mathbb{E}_{x\sim\mathcal{D}_f}\left[r(x,s)\right].
\end{equation}
We omit $E$ from the notation because the executor is fixed. The selector should keep the skill with the higher utility. This utility is unknown at deployment. We therefore learn the relative order of skills from measurements made before deployment. In this paper, cross-task transfer means transfer between different tasks of the same family. It does not mean transfer between families.
\section{Method}
\label{sec:method}
\begin{figure*}[t]
\centering
\includegraphics[width=\textwidth]{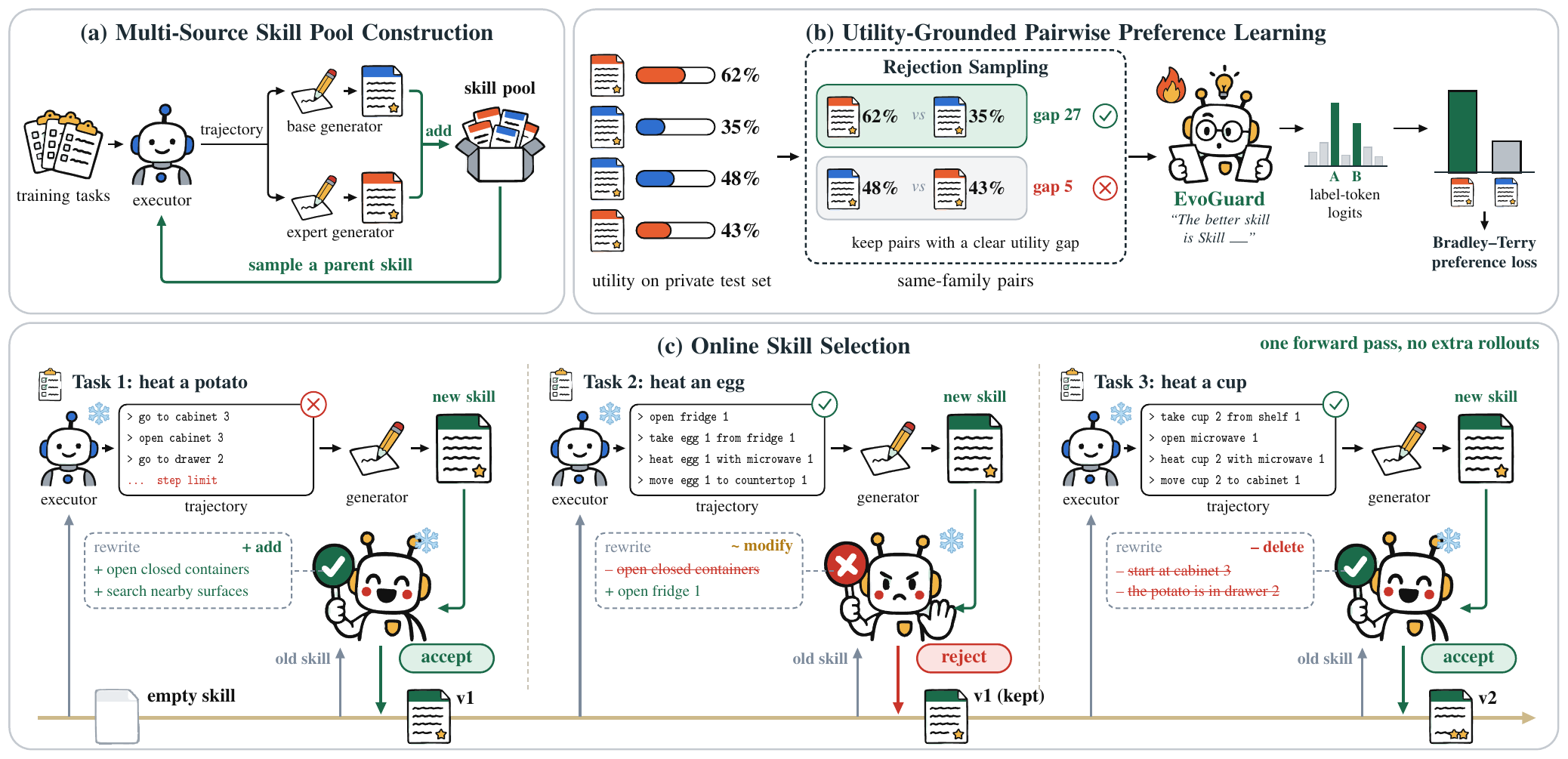}
\caption{Overview of \method{}. \textbf{(a)} A rewrite loop on training tasks fills a skill pool from two sources, a base generator and an expert generator. \textbf{(b)} The utility of every skill is estimated on a private test set. Rejection sampling keeps same-family pairs with a clear utility gap to train a pairwise selector. \textbf{(c)} At deployment the frozen selector accepts or rejects each rewrite in one forward pass.}
\label{fig:overview}
\end{figure*}

Figure~\ref{fig:overview} shows the workflow of \method{}. Before deployment, we first construct a multi-source skill pool with the rewrite loop. We then estimate the utility of every skill on a private test set. Finally, we train a pairwise selector on pairs of scored skills. At deployment the frozen selector decides after every task whether the new skill replaces the current one. We drop the task index $t$ in this section.

\subsection{Single-Trajectory Skill Rewriting}
\label{sec:rewrite}
A skill is a \texttt{SKILL.md} document. It has YAML metadata and four Markdown sections. The sections are \emph{When to use}, \emph{Workflow}, \emph{Decision rules}, and \emph{Recovery and pitfalls}. After each task, the generator $G$ rewrites the whole document. Its inputs are the current skill $s$, the trajectory $\tau$ of actions and observations, and the outcome $r$:
\begin{equation}
\tilde{s}=G(s,\tau,r).
\end{equation}
The format check $v(\tilde{s})$ is purely structural. It requires the family-specific name and a nonempty description. It also requires each of the four sections to appear exactly once with nonempty content. The check does not test whether the skill helps. A rewrite with $v(\tilde{s})=0$ is discarded, and the current skill stays.

\subsection{Multi-Source Skill Pool Construction}
\label{sec:pool}
For each family $f$, we split the training tasks into two parts. A small part serves as the private test set $\mathcal{H}_f$. The other part is used to build the skill pool $\mathcal{P}_f$. The pool is filled by the rewrite loop itself. The first few tasks run without a skill, and their valid rewrites initialize the pool. For every later task $x$, we draw a parent skill $s$ uniformly from the pool. The executor attempts $x$ with $s$ and returns $(\tau,r)=E(x,s)$. The generator then rewrites $s$. The new skill joins the pool if it passes the format check:
\begin{equation}
\mathcal{P}_f\leftarrow\mathcal{P}_f\cup\{G(s,\tau,r)\}.
\end{equation}
We record $s$ as the parent of the new skill.

\paragraph{Two skill sources.}
The pool should contain both good and poor skills with clear utility gaps between them. Such gaps give the selector informative training pairs. Skills from a single generator tend to cover only a narrow quality range. We therefore fill the pool from two sources. The first source is the base generator $G$ itself. The second source is an expert generator, a stronger model that writes skills of higher quality. Section~\ref{sec:setup} gives the expert generator of each benchmark.

\subsection{Skill Utility Estimation}
\label{sec:degrade}
We evaluate every valid skill once on each task of $\mathcal{H}_f$. We also evaluate the empty skill of each family. The empirical utility of a skill is its mean outcome on the private test set:
\begin{equation}
\hat{u}_f(s)=\frac{1}{|\mathcal{H}_f|}\sum_{x\in\mathcal{H}_f}r(x,s).
\label{eq:utility}
\end{equation}
The tasks in $\mathcal{H}_f$ are never used to generate skills. They are also disjoint from the online evaluation tasks.

\subsection{Pairwise Preference Learning}
\label{sec:train}
\paragraph{Skill-level split.}
We split the scored skills into training skills and validation skills. The split is stratified by family and source. The empty skills stay on the training side. A validation skill never appears in a training pair.

\paragraph{Rejection sampling of pairs.}
We draw two training skills $s_i$ and $s_j$ of the same family. We keep the pair only if their utilities are clearly apart:
\begin{equation}
\left|\hat{u}_f(s_i)-\hat{u}_f(s_j)\right|\ge\delta.
\label{eq:reject}
\end{equation}
Each utility is estimated from a finite set of tasks. Pairs with a smaller gap therefore have unreliable labels, and we reject them. No unordered pair is kept twice. We place the two skills at positions A and B in random order. The label $y\in\{A,B\}$ is the position of the skill with the higher utility. A pair does not need to consist of a parent and its child.

\paragraph{Pairwise preference objective.}
We cast the comparison as pairwise preference learning over two label tokens. The selector is a cross-encoder that reads both skills in one prompt. The prompt names the family and describes its generic task. It then presents the two texts as Skill A and Skill B. The prompt ends with the assistant prefix \texttt{The better skill is Skill}. The answer is thus the next token. Both \texttt{\textvisiblespace A} and \texttt{\textvisiblespace B} are single tokens in the vocabulary. Let $\mathbf{h}_\theta\in\mathbb{R}^d$ be the final-layer hidden state at the last prompt position. Here $\theta$ denotes the trainable parameters. Let $\mathbf{w}_A$ and $\mathbf{w}_B$ be the rows of the unembedding matrix for the two label tokens. The selector uses only the logits of these two tokens instead of the full next-token distribution:
\begin{equation}
z_A=\mathbf{w}_A^{\top}\mathbf{h}_\theta,\qquad z_B=\mathbf{w}_B^{\top}\mathbf{h}_\theta.
\end{equation}
We renormalize over the two tokens to obtain the probability that the skill at position $y$ is better. We then minimize its negative log-likelihood:
\begin{equation}
\mathcal{L}(\theta)=-\log\frac{\exp z_y}{\exp z_A+\exp z_B}.
\label{eq:loss}
\end{equation}
This probability equals $\sigma(z_y-z_{\bar{y}})$, where $\bar{y}$ is the other position. The loss thus depends only on the margin between the two logits and has the form of a Bradley--Terry preference loss \citep{bradley1952rank}. 

\subsection{Online Skill Selection}
\label{sec:online}
At deployment every family starts from the empty skill. Consider a task of family $f$ whose rewrite $\tilde{s}$ passes the format check. We place the current skill $s_f$ and $\tilde{s}$ at positions A and B in random order, as in training. The selector is then queried once. It returns the position with the larger logit. The skill at that position becomes the family skill:
\begin{equation}
a^\star=
\begin{cases}
A & \text{if } z_A\ge z_B,\\
B & \text{otherwise},
\end{cases}
\qquad s_f\leftarrow s_{a^\star}.
\label{eq:accept}
\end{equation}
This rule implements the selector $C$ in Eq.~\eqref{eq:select}. The prompt is the same as in training. One decision costs one forward pass. It generates no tokens and runs no extra rollouts. Neither the selector nor the executor is updated online. Appendix~\ref{sec:prompts} gives the prompts, Appendix~\ref{sec:inference} describes how the selector is served, and Appendix~\ref{sec:loop} gives the full online loop.
\section{Experimental Setup}
\label{sec:setup}
\subsection{Benchmarks}
We evaluate on four benchmarks and use their test sets for online evaluation. ALFWorld \citep{shridhar2021alfworld} is a text-based household environment. An agent completes tasks such as finding, heating, or cleaning objects through text commands. Its tasks fall into six families, namely Pick, Look, Clean, Heat, Cool, and Pick2. We use its seen and unseen test splits with 140 and 134 tasks. ScienceWorld \citep{wang2022scienceworld} is a text-based environment for elementary science experiments. Each of its 30 task types forms a family. We draw at most 30 test tasks per type, which gives 651 test tasks. LiveCodeBench \citep{jain2024livecodebench} collects competitive programming problems over time. We use the 175 problems added in its sixth release as the test set. Codex with GPT-6.1-sol assigns each problem to one of nine algorithmic categories (Appendix~\ref{sec:livecode}). These categories serve as families. WebShop \citep{yao2022webshop} simulates online shopping. An agent searches a store and buys a product that matches an instruction. Its five product categories serve as families, and we use its 500 test tasks.

\subsection{Evaluation Metrics}
On ALFWorld, we report the success rate of each family and of the whole split. On ScienceWorld, we report the success rate and the mean environment score. The score ranges from 0 to 100 and also credits partial progress. On LiveCodeBench, we report the pass rate. A problem counts as passed if the generated program passes all hidden tests. On WebShop, we report the success rate and the mean purchase score. The score rates the purchased product on its type, attributes, options, and price.

\subsection{Baselines}
\noindent\textbf{\textit{No-Skill}} runs the executor with an empty skill on every task.\par
\noindent\textbf{\textit{Accept-All}} adopts every valid rewrite.\par
\noindent\textbf{\textit{Random}} samples a skill uniformly from all valid rewrites of the family so far, including the empty skill.\par
\noindent\textbf{\textit{LLM-Judge}} gives the selector prompt to the untrained backbone and lets it generate A or B.\par
\noindent\textbf{\textit{LLM-Judge (Think)}} does the same with thinking enabled, so the model reasons before it answers.\par
\noindent\textbf{\textit{LLM-Judge (Logit)}} applies our assistant prefix and our A/B logit rule to the untrained backbone.\par

\subsection{Implementation Details}
\paragraph{Training.}
To build pairs with clear utility gaps, we sample skills from generators of different quality. Skills from different generators follow different distributions and cover different quality levels. We use Qwen3-4B \citep{yang2025qwen3} as the generator. However, its skills are of limited quality. Its skills alone rarely form pairs with clear utility gaps. We therefore add a second source of skills. On ALFWorld, ScienceWorld, and WebShop, the second source is an expert generator. We train it from Qwen3-4B with GRPO on the training tasks. On LiveCodeBench, the second source is GPT-6.1-sol. The second source samples more high-quality skills. Together with the base skills, it yields pairs with clear utility gaps.

All experiments run on a server with four NVIDIA H200 GPUs. The selector backbone stays frozen, and we train LoRA adapters \citep{hu2021lora} on all attention and feed-forward projections. The adapters have rank 32 and scaling factor 64. We use AdamW with a learning rate of $10^{-4}$ and a cosine schedule. We train for one epoch and use the final checkpoint. Appendix~\ref{sec:seltrain} gives the remaining details.

\paragraph{Evaluation.}
The online evaluation samples at temperature 1.0. An episode has at most 30 actions on ALFWorld, ScienceWorld, and WebShop. Each setting is run three times with different sampling seeds. We report the mean over the three runs. WebShop results are reported in Appendix~\ref{sec:webshop}.
\section{Experimental Results}
\label{sec:results}
\subsection{Main Performance}
\label{sec:main}
Table~\ref{tab:main} reports online success on ALFWorld. We train the selector once. We then evaluate the same selector with executors built on three backbones, namely Qwen3-4B, Qwen3-8B, and Qwen3-32B.

\begin{table*}[t]
\centering
\footnotesize
\setlength{\tabcolsep}{3.2pt}
\resizebox{\textwidth}{!}{\begin{tabular}{lcccccc>{\columncolor{allcol}}cp{3pt}cccccc>{\columncolor{allcol}}c}
\toprule
& \multicolumn{7}{c}{\textbf{Seen}} & & \multicolumn{7}{c}{\textbf{Unseen}} \\
\cmidrule(lr){2-8}\cmidrule(lr){10-16}
\textbf{Method} & \textbf{Pick} & \textbf{Look} & \textbf{Clean} & \textbf{Heat} & \textbf{Cool} & \textbf{Pick2} & \textbf{All} & & \textbf{Pick} & \textbf{Look} & \textbf{Clean} & \textbf{Heat} & \textbf{Cool} & \textbf{Pick2} & \textbf{All} \\
& (35) & (13) & (27) & (16) & (25) & (24) & (140) & & (24) & (18) & (31) & (23) & (21) & (17) & (134) \\
\midrule
\multicolumn{16}{l}{\textit{Executor: Qwen3-4B}} \\
No-Skill & $82.9$ & $10.3$ & $23.5$ & $10.4$ & $1.3$ & $48.6$ & $36.0$ & & $81.9$ & $\mathbf{29.6}$ & $29.0$ & $13.0$ & $11.1$ & $23.5$ & $32.3$ \\
Accept-All & $90.5$ & $20.5$ & $27.2$ & $29.2$ & $14.7$ & $30.6$ & $41.0$ & & $\mathbf{86.1}$ & $11.1$ & $\mathbf{78.5}$ & $30.4$ & $54.0$ & $19.6$ & $51.2$ \\
Random & $87.6$ & $12.8$ & $37.0$ & $33.3$ & $16.0$ & $30.6$ & $42.1$ & & $75.0$ & $16.7$ & $58.1$ & $30.4$ & $44.4$ & $25.5$ & $44.5$ \\
LLM-Judge & $82.9$ & $5.1$ & $35.8$ & $25.0$ & $29.3$ & $38.9$ & $42.9$ & & $77.8$ & $20.4$ & $\mathbf{78.5}$ & $21.7$ & $60.3$ & $23.5$ & $51.0$ \\
LLM-Judge (Think) & $\mathbf{91.4}$ & $15.4$ & $42.0$ & $39.6$ & $24.0$ & $18.1$ & $44.3$ & & $76.4$ & $1.9$ & $67.7$ & $47.8$ & $50.8$ & $19.6$ & $48.3$ \\
LLM-Judge (Logit) & $85.7$ & $12.8$ & $45.7$ & $22.9$ & $22.7$ & $38.9$ & $44.8$ & & $84.7$ & $11.1$ & $67.7$ & $20.3$ & $57.1$ & $23.5$ & $47.8$ \\
\rowcolor{oursrow}\method{} & $87.6$ & $\mathbf{33.3}$ & $\mathbf{46.9}$ & $\mathbf{54.2}$ & $\mathbf{37.3}$ & $\mathbf{51.4}$ & \cellcolor{oursall}$\mathbf{55.7}$ & & $84.7$ & $24.1$ & $76.3$ & $\mathbf{68.1}$ & $\mathbf{65.1}$ & $\mathbf{35.3}$ & \cellcolor{oursall}$\mathbf{62.4}$ \\
\addlinespace[3pt]
\multicolumn{16}{l}{\textit{Executor: Qwen3-8B}} \\
No-Skill & $85.7$ & $23.1$ & $34.6$ & $8.3$ & $8.0$ & $\mathbf{61.1}$ & $43.1$ & & $86.1$ & $22.2$ & $32.3$ & $5.8$ & $12.7$ & $\mathbf{51.0}$ & $35.3$ \\
Accept-All & $\mathbf{89.5}$ & $20.5$ & $37.0$ & $2.1$ & $30.7$ & $41.7$ & $44.3$ & & $86.1$ & $22.2$ & $68.8$ & $2.9$ & $39.7$ & $35.3$ & $45.5$ \\
Random & $88.6$ & $\mathbf{35.9}$ & $35.8$ & $4.2$ & $17.3$ & $47.2$ & $44.0$ & & $\mathbf{87.5}$ & $24.1$ & $61.3$ & $8.7$ & $50.8$ & $35.3$ & $47.0$ \\
LLM-Judge & $79.0$ & $23.1$ & $37.0$ & $8.3$ & $34.7$ & $40.3$ & $43.1$ & & $\mathbf{87.5}$ & $29.6$ & $72.0$ & $2.9$ & $23.8$ & $35.3$ & $45.0$ \\
LLM-Judge (Think) & $81.9$ & $20.5$ & $42.0$ & $4.2$ & $10.7$ & $56.9$ & $42.6$ & & $79.2$ & $\mathbf{33.3}$ & $73.1$ & $1.4$ & $50.8$ & $25.5$ & $47.0$ \\
LLM-Judge (Logit) & $85.7$ & $20.5$ & $28.4$ & $2.1$ & $32.0$ & $37.5$ & $41.2$ & & $84.7$ & $\mathbf{33.3}$ & $\mathbf{77.4}$ & $1.4$ & $39.7$ & $33.3$ & $48.3$ \\
\rowcolor{oursrow}\method{} & $88.6$ & $30.8$ & $\mathbf{51.9}$ & $\mathbf{16.7}$ & $\mathbf{44.0}$ & $50.0$ & \cellcolor{oursall}$\mathbf{53.3}$ & & $86.1$ & $31.5$ & $75.3$ & $\mathbf{20.3}$ & $\mathbf{52.4}$ & $47.1$ & \cellcolor{oursall}$\mathbf{54.7}$ \\
\addlinespace[3pt]
\multicolumn{16}{l}{\textit{Executor: Qwen3-32B}} \\
No-Skill & $89.5$ & $66.7$ & $33.3$ & $29.2$ & $22.7$ & $54.2$ & $51.7$ & & $88.9$ & $50.0$ & $49.5$ & $\mathbf{50.7}$ & $57.1$ & $51.0$ & $58.2$ \\
Accept-All & $90.5$ & $71.8$ & $38.3$ & $10.4$ & $42.7$ & $47.2$ & $53.6$ & & $86.1$ & $57.4$ & $80.6$ & $15.9$ & $76.2$ & $37.3$ & $61.2$ \\
Random & $88.6$ & $76.9$ & $\mathbf{56.8}$ & $29.2$ & $30.7$ & $59.7$ & $59.3$ & & $90.3$ & $63.0$ & $68.8$ & $21.7$ & $\mathbf{85.7}$ & $51.0$ & $64.2$ \\
LLM-Judge & $87.6$ & $\mathbf{84.6}$ & $33.3$ & $14.6$ & $40.0$ & $43.1$ & $52.4$ & & $86.1$ & $\mathbf{79.6}$ & $\mathbf{88.2}$ & $15.9$ & $81.0$ & $\mathbf{52.9}$ & $68.7$ \\
LLM-Judge (Think) & $89.5$ & $74.4$ & $44.4$ & $14.6$ & $45.3$ & $62.5$ & $58.3$ & & $86.1$ & $66.7$ & $\mathbf{88.2}$ & $7.2$ & $66.7$ & $39.2$ & $61.4$ \\
LLM-Judge (Logit) & $88.6$ & $74.4$ & $35.8$ & $8.3$ & $21.3$ & $52.8$ & $49.8$ & & $88.9$ & $70.4$ & $81.7$ & $7.2$ & $77.8$ & $49.0$ & $63.9$ \\
\rowcolor{oursrow}\method{} & $\mathbf{96.2}$ & $76.9$ & $51.9$ & $\mathbf{33.3}$ & $\mathbf{52.0}$ & $\mathbf{72.2}$ & \cellcolor{oursall}$\mathbf{66.7}$ & & $\mathbf{91.7}$ & $\mathbf{79.6}$ & $87.1$ & $33.3$ & $77.8$ & $51.0$ & \cellcolor{oursall}$\mathbf{71.6}$ \\
\bottomrule
\end{tabular}
}
\caption{Online success rate (\%, $\uparrow$) on ALFWorld by task family for three executors. Each value is the mean of three runs. Bold marks the best value in each column for each executor.}
\label{tab:main}
\end{table*}

\paragraph{Overall success.}
\method{} achieves the best overall success with every executor. With the 4B executor, it reaches 55.7\% on seen and 62.4\% on unseen tasks. The strongest baselines reach 44.8\% and 51.2\%. With the 8B executor, \method{} reaches 53.3\% and 54.7\%. The strongest baselines reach 44.3\% and 48.3\%. Averaged over these four settings, \method{} reaches 56.5\%. Accept-All and the three LLM-Judge baselines all average about 45.5\%. The executor alone averages 36.7\%.

\paragraph{An executor without training labels.}
No training label comes from Qwen3-32B. Its rows therefore test transfer to an unseen executor. \method{} reaches 66.7\% on seen and 71.6\% on unseen tasks. The strongest baselines reach 59.3\% and 68.7\%. The executor alone reaches 51.7\% and 58.2\%.

\paragraph{Effect of training.}
LLM-Judge (Logit) uses the same prompt and the same decision rule as \method{}. The comparison between the two thus isolates the effect of training. \method{} improves over LLM-Judge (Logit) by 6.4\% to 16.9\% in overall success.

\paragraph{Per-family results.}
The gains differ across task families. They are largest on families that need multi-step procedures. With the 4B executor on seen tasks, \method{} raises Heat from 10.4\% without skills to 54.2\%. It raises Cool from 1.3\% to 37.3\%. On Pick, every method already reaches at least 75\%, which leaves little room for skills. On a few families, skills even hurt. On Pick2 with the 8B executor, the executor alone reaches 61.1\% on seen tasks, while \method{} reaches 50.0\%. On Heat with the 32B executor, the executor alone reaches 50.7\% on unseen tasks, while \method{} reaches 33.3\%. The selector can only choose among the rewrites it receives. It cannot repair a family where written skills do not help.

\paragraph{Acceptance behavior.}
Table~\ref{tab:acceptance} reports how each method treats the rewrites. Almost all rewrites pass the format check. Only LLM-Judge and LLM-Judge (Think) see a few invalid ones. \method{} accepts only 42.3\% and 43.2\% of the valid rewrites. The LLM-Judge baselines accept 68.4\% to 72.0\%. The skill in use is also shorter under \method{}. It averages 361 to 371 tokens, against 528 to 567 tokens under Accept-All and the LLM-Judge baselines. Yet \method{} achieves the best overall success with these shorter skills. This reveals that \method{} tends to keep content that is general and useful, rather than content that is specific to individual cases.

\begin{table}[t]
\centering
\footnotesize
\setlength{\tabcolsep}{3pt}
\resizebox{\columnwidth}{!}{\begin{tabular}{lcccccc}
\toprule
 & \multicolumn{3}{c}{\textbf{Qwen3-4B}} & \multicolumn{3}{c}{\textbf{Qwen3-8B}} \\
\cmidrule(lr){2-4}\cmidrule(lr){5-7}
\textbf{Method} & Valid & Accepted & Tokens & Valid & Accepted & Tokens \\
\midrule
Accept-All & 100.0 & 100.0 & 553 & 100.0 & 100.0 & 567 \\
Random & 100.0 & 12.8 & 303 & 100.0 & 13.0 & 307 \\
LLM-Judge & 99.8 & 68.9 & 541 & 99.9 & 70.9 & 536 \\
LLM-Judge (Think) & 98.4 & 68.4 & 565 & 99.1 & 68.8 & 548 \\
LLM-Judge (Logit) & 100.0 & 69.0 & 543 & 100.0 & 72.0 & 528 \\
\rowcolor{oursrow}\method{} & 100.0 & 42.3 & 371 & 100.0 & 43.2 & 361 \\
\bottomrule
\end{tabular}
}
\caption{Acceptance statistics of the runs in Table~\ref{tab:main}. Valid is the share of rewrites that pass the format check (\%). Accepted is the share of valid rewrites that are adopted (\%). Tokens is the mean length of the skill in use.}
\label{tab:acceptance}
\end{table}

\subsection{Performance on Other Benchmarks}
\label{sec:envs}
We also report results on ScienceWorld and LiveCodeBench in Table~\ref{tab:envs}.

\begin{table}[t]
\centering
\footnotesize
\setlength{\tabcolsep}{3pt}
\resizebox{\columnwidth}{!}{\begin{tabular}{lccc}
\toprule
& \multicolumn{2}{c}{\textbf{ScienceWorld}} & \textbf{LiveCodeBench} \\
\cmidrule(lr){2-3}\cmidrule(lr){4-4}
\textbf{Method} & Success & Score & Pass Rate \\
\midrule
No-Skill & $10.2$ & $20.6$ & $26.1$ \\
Accept-All & $9.4$ & $21.2$ & $25.7$ \\
Random & $10.0$ & $20.7$ & $25.7$ \\
LLM-Judge & $10.0$ & $21.6$ & $26.7$ \\
LLM-Judge (Think) & $10.8$ & $21.3$ & $25.3$ \\
LLM-Judge (Logit) & $9.6$ & $22.0$ & $27.0$ \\
\rowcolor{oursrow}\method{} & $\mathbf{15.3}$ & $\mathbf{27.3}$ & $\mathbf{28.4}$ \\
\bottomrule
\end{tabular}
}
\caption{Online results on ScienceWorld and LiveCodeBench (\%). ScienceWorld reports the success rate and the mean score. LiveCodeBench reports the pass rate. Bold marks the best value in each column.}
\label{tab:envs}
\end{table}

\paragraph{ScienceWorld.}
\method{} reaches 15.3\% success and a score of 27.3. Without skills, the executor reaches 10.2\% and 20.6. The other methods that use skills reach between 9.4\% and 10.8\% success. None of them is clearly better than using no skill. Accept-All even falls below it.

\paragraph{LiveCodeBench.}
\method{} achieves the best pass rate of 28.4\%. This is 2.3\% higher than running without skills. In contrast, Accept-All and Random both drop slightly to 25.7\%. The LLM-Judge baselines reach between 25.3\% and 27.0\%, which is close to running without skills.

\paragraph{Analysis.}
The two benchmarks show the same pattern. The other methods that use skills stay close to the executor without skills, and Accept-All falls below it on both benchmarks. These methods cannot evolve the skill in a stable way. Good rewrites are mixed with harmful ones, and the gains of the former are cancelled by the latter. \method{} filters out the harmful rewrites before they replace the current skill. The skill can therefore improve over the task stream instead of drifting. These results show that \method{} mitigates skill degradation during evolution, and that this benefit is not limited to ALFWorld.

\subsection{Training Performance}
\label{sec:accuracy}
\begin{figure}[t]
\centering
\includegraphics[width=\columnwidth]{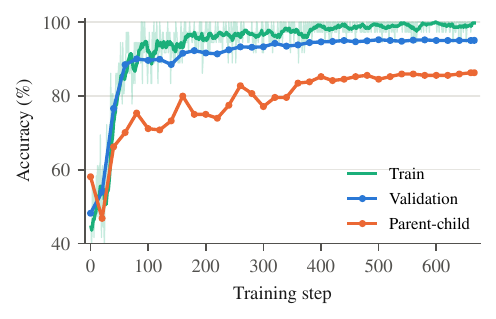}
\caption{Selector accuracy during training on the training batches (Train), on validation pairs (Validation), and on parent-child validation pairs (Parent-child).}
\label{fig:training}
\end{figure}

Figure~\ref{fig:training} tracks the selector during training. On validation pairs, the untrained Qwen3-4B reaches 48.2\%. This is no better than chance, because it answers A for 72\% of the prompts. Accuracy rises to 89.6\% after 100 steps. It reaches 95.0\% at the final step. Parent-child pairs match the comparison of the online loop. On these pairs, accuracy rises from 58.1\% to 86.3\%.

\subsection{Position Stability}
\label{sec:position}
\begin{figure}[t]
\centering
\includegraphics[width=\columnwidth]{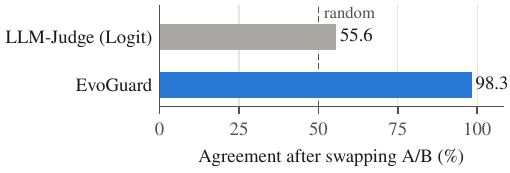}
\caption{Share of validation pairs for which the selector keeps the same skill after the two skills swap positions. The dashed line marks a random choice.}
\label{fig:position}
\end{figure}

Pairwise LLM judges are known to suffer from position bias \citep{wang2023notfair,zheng2023judging}. We test this bias on the validation pairs. For each pair, we swap the positions of the two skills and check whether the selector keeps the same skill. Figure~\ref{fig:position} reports this agreement rate. The untrained LLM-Judge (Logit) agrees on only 55.6\% of the pairs. This is close to the 50\% of a random choice. The untrained backbone thus cannot judge reliably which of the two skills is better. Our training places the two skills in random order and optimizes the two label logits. This calibrates the logits and makes the decision nearly invariant to the order of the inputs. \method{} agrees on 98.3\% of the pairs.

\subsection{Ablation Study}
\label{sec:ablation}
Table~\ref{tab:ablation} removes one component of the training data at a time. Both variants reduce the average success by more than 4\%. \emph{w/o expert-written pool} removes the expert-written skills from the pool. The pool then lacks high-quality skills, because the base generator writes skills of limited quality. These skills are also generated by the same backbone as the selector, so their distribution is close to that of the selector backbone itself. The resulting pairs are difficult to distinguish and provide only a weak training signal. \emph{w/o rejection sampling} draws pairs at random and discards only ties. As a result, 44\% of the training pairs differ by less than 10\% in utility, and their labels are dominated by estimation noise. Rejection sampling retains only pairs with a clear utility gap, which allows the selector to learn the features that separate good skills from poor ones. Both components are thus needed to obtain informative training pairs.

\begin{table}[t]
\centering
\footnotesize
\setlength{\tabcolsep}{3pt}
\resizebox{\columnwidth}{!}{\begin{tabular}{lccccc}
\toprule
& \multicolumn{2}{c}{\textbf{Qwen3-4B}} & \multicolumn{2}{c}{\textbf{Qwen3-8B}} & \\
\cmidrule(lr){2-3}\cmidrule(lr){4-5}
\textbf{Variant} & Seen & Unseen & Seen & Unseen & \textbf{Avg.} \\
\midrule
\method{} & $55.7$ & $62.4$ & $53.3$ & $54.7$ & 56.5 \\
\ \ w/o expert-written pool & $47.1$ & $58.2$ & $48.3$ & $54.5$ & 52.0 \\
\ \ w/o rejection sampling & $49.5$ & $57.0$ & $48.3$ & $54.7$ & 52.4 \\
\bottomrule
\end{tabular}
}
\caption{Ablation study of \method{}, reported as overall online success (\%).}
\label{tab:ablation}
\end{table}

\section{Conclusion}
Online skill evolution lets a deployed agent improve its skill after every task. However, each rewrite is based on a single trajectory and can degrade the skill. Adopting every rewrite lets the skill drift, while validating every update is too costly. We propose \method{}, a pairwise skill selector for skill selection. We first build a dataset of skills with utility labels through iterative rollouts and skill rewriting. Each label is the empirical utility of the skill on a private test set. Rejection sampling then keeps high-confidence pairs with clear utility gaps as preference supervision. We cast the comparison between the current skill and the new one as pairwise preference learning. The selector takes only the logits of the two label tokens and renormalizes over them. It aligns this conditional distribution with the preference labels. At deployment, each decision needs a single forward pass and generates no tokens. Its cost and latency are therefore low. Experiments on two agent benchmarks and one code benchmark confirm the effectiveness of \method{}. On ALFWorld, it reaches 56.5\% average success against 45.5\% for adopting every rewrite. On ScienceWorld, it raises success from 10.2\% to 15.3\%. On LiveCodeBench, it achieves the best pass rate of 28.4\%.

\section*{Limitations}
Our experiments cover text-based agent environments and one code generation benchmark. Extending \method{} to multimodal or embodied settings is left for future work. All executors come from one model family. Other model families may rank skills somewhat differently. Building and scoring the skill pool requires offline computation before deployment. This cost is paid once and is shared by all later decisions. We train one selector per benchmark. A single selector that transfers across domains would further reduce the setup cost. The selector compares the current skill with one new skill at a time. Richer update rules, such as keeping several versions or reverting to an earlier one, may bring further gains. Finally, the selector can only choose among the rewrites it receives. Its benefit is therefore smaller in settings where skills offer little help.
\section*{Ethical Considerations}
Our experiments use public benchmarks in simulated environments and model-generated trajectories. They involve no human subjects and no private user data. The selector is supervised by task success, which does not guarantee that a skill is safe outside these environments. A reusable skill may also carry incorrect instructions to later tasks. Real deployments should therefore add safety checks suited to their domain and keep a record of skill versions. We use all benchmarks and models in accordance with their licenses. Generative AI assisted with literature search and with drafting and editing this manuscript.
\bibliography{references}
\appendix
\raggedbottom
\section{Additional Implementation Details}
\label{sec:appendix}
This appendix gives the problem categories of LiveCodeBench, the prompts, the training of the selector, its use at inference, and the full online loop.

\subsection{LiveCodeBench Problem Categories}
\label{sec:livecode}
LiveCodeBench provides no grouping of its problems into task families. We therefore group the problems by the kind of algorithm they need, so that problems of one category can share a skill. We use the 879 problems released up to version 5 that have sample tests as the training set and the 175 problems added in version 6 as the test set \citep{jain2024livecodebench}.

\paragraph{Labeling procedure.}
We label all 1{,}054 problems with GPT-6.1-sol through the Codex command-line interface. Every call runs in an empty directory with a read-only sandbox. The prompt starts with the instruction to answer from the given text only, to run no commands, and to reply with one JSON value. The model sees the title, the statement, and the starter code of a problem. The hidden tests are never shown to it. The training and the test problems go through the same procedure, which has two steps.

In the first step, the model tags the problems in batches of 20. The prompt is followed by the problems:
\par\vspace{4pt}\noindent\begin{minipage}{\columnwidth}
\begin{lstlisting}
Below are {N} competitive programming problems. For each one, name the algorithmic topics a solver needs for the intended solution: 1 to 3 short lower-case tags (for example "dynamic programming", "greedy", "graph traversal", "simulation"), the most specific that still describe a family of problems, and which one of them is the primary topic.

Reply with a JSON array with one object per problem, in the same order: {"id": "<problem id>", "tags": ["<tag>", ...], "primary": "<one of the tags>"}
\end{lstlisting}

\end{minipage}\par\vspace{4pt}
An answer is kept only if the primary topic is one of the listed tags. This step yields 206 distinct primary tags.

In the second step, one call merges these tags into categories. The prompt is followed by every tag with its number of problems:
\par\vspace{4pt}\noindent\begin{minipage}{\columnwidth}
\begin{lstlisting}
{N} competitive programming problems (LeetCode and AtCoder) were each given one primary topic tag. Below is every tag with its number of problems. Merge the tags into 8 to 12 categories so that problems of one category are solved with a shared set of techniques and a note written for the category could help on each of its problems. Every tag goes to exactly one category; no category should hold fewer than about 4% of the problems.

Reply with a JSON object: {"categories": [{"name": "<lower-case-kebab-case id>", "title": "<short title>", "description": "<one sentence in the imperative saying what such problems ask for, e.g. Find an optimal value by filling a table over subproblems.>"}, ...], "tags": {"<tag>": "<category name>", ...}}
\end{lstlisting}

\end{minipage}\par\vspace{4pt}
We accept the answer only if it has 8 to 12 categories and assigns every tag to one of them. Each problem then belongs to the category of its primary tag. The one-sentence description of a category serves as its family description in the selector prompt.

\begin{figure}[!htb]
\centering
\includegraphics[width=\columnwidth]{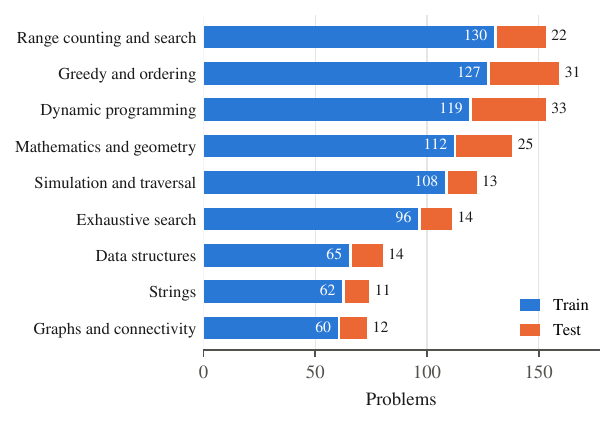}
\caption{Number of LiveCodeBench problems per category in the training and the test set.}
\label{fig:livecodecategories}
\end{figure}

\paragraph{Category distribution.}
The procedure yields nine categories. Figure~\ref{fig:livecodecategories} shows their sizes in the training and the test set. The categories are reasonably balanced. The largest one, Greedy and ordering, holds 158 problems (15.0\%), and the smallest one, Graphs and connectivity, holds 72 (6.8\%). No category falls below the 4\% floor requested in the prompt. The four largest categories are Greedy and ordering, Dynamic programming, Range counting and search, and Mathematics and geometry. Together they cover 56.8\% of the problems. The test set follows a similar distribution. Every category has at least 11 test problems, so each skill is updated several times during the online evaluation.

\subsection{Prompts}
\label{sec:prompts}
Figure~\ref{fig:prompts} shows the two prompts of \method{} on ALFWorld. The selector prompt names the task family and gives its generic description. It then presents the two skills within \texttt{<skill\_a>} and \texttt{<skill\_b>} delimiters and asks which one should be kept. An empty skill is shown as \texttt{(empty skill)}. \method{} and LLM-Judge (Logit) append the assistant prefix \texttt{The better skill is Skill}, so that the answer is the next token. LLM-Judge and LLM-Judge (Think) use no prefix and generate their answer. The rewrite prompt gives the generator the current skill and the latest attempt. It asks for the complete updated skill in the format of \S\ref{sec:rewrite}.

\begin{figure*}[!t]
\centering
\begin{minipage}[t]{\textwidth}
\textbf{\small (a) Selector prompt}
\begin{lstlisting}
An agent solves ALFWorld household tasks of type "{family}": {family description}
It reads one reusable skill at the start of every task of this type. Keep the better of the two skills below, the one with which the agent will succeed more often on future tasks of this type.

Skill A:
<skill_a>
{skill A, or "(empty skill)"}
</skill_a>

Skill B:
<skill_b>
{skill B, or "(empty skill)"}
</skill_b>

Which skill should be kept? Answer with A or B.
\end{lstlisting}
\vspace{2pt}
{\footnotesize Assistant prefix of \method{}:}
\begin{lstlisting}
The better skill is Skill
\end{lstlisting}
\end{minipage}

\vspace{6pt}
\begin{minipage}[t]{\textwidth}
\textbf{\small (b) Rewrite prompt}
\begin{lstlisting}
You maintain a reusable skill for an agent that solves ALFWorld household tasks of type "{family}": {family description}
The agent reads the skill at the start of every task of this type.

Current skill:
{current skill, or "(none yet)"}

The agent just attempted one task with the current skill:
{task, outcome, and the action and observation of every step}

Write the updated skill for future tasks of this type, using what this attempt shows.
Available commands: look; inventory; go to {receptacle}; open {receptacle}; close {receptacle}; take {object} from {receptacle}; move {object} to {receptacle}; examine {target}; use {lamp}; heat {object} with {microwave}; clean {object} with {sink}; cool {object} with {fridge}; slice {object} with {knife}.

Output the complete skill as a standard SKILL.md document inside <skill> and </skill>, at most 1500 tokens, and nothing else:
<skill>
---
name: alfworld-{family}
description: <what the skill does and when to use it>
---
# ALFWorld {family}
## When to use
...
## Workflow
...
## Decision rules
...
## Recovery and pitfalls
...
</skill>
\end{lstlisting}
\end{minipage}
\caption{The two prompts of \method{} on ALFWorld. Braces mark inserted content. (a) The selector reads the family description and the two skills. It compares the next-token logits of A and B after the assistant prefix. (b) The generator rewrites the skill after every task. In the prompt the four section headings are separated by blank lines.}
\label{fig:prompts}
\end{figure*}

\paragraph{Prompts on the other benchmarks.}
The other benchmarks reuse the two prompts of Figure~\ref{fig:prompts}. Only the parts listed below change.

\textbf{ScienceWorld.} The domain wording becomes ``ScienceWorld science-experiment tasks of type'', and the skill is named \texttt{scienceworld-\{family\}}. The command list of the rewrite prompt is:
\par\vspace{4pt}\noindent\begin{minipage}{\columnwidth}
\begin{lstlisting}
look around; look at {object}; look in {object}; go {location}; open {object}; close {object}; pick up {object}; put down {object}; move {object} to {object}; pour {object} in {object}; dunk {object} in {object}; mix {object}; activate {object}; deactivate {object}; use {object} on {object}; connect {object} to {object}; disconnect {object}; read {object}; focus on {object}; eat {object}; flush {object}; inventory; task; wait; wait1
\end{lstlisting}

\end{minipage}\par\vspace{4pt}
The executor prompt, which is the same for every method, also carries the following advice on the action format:
\par\vspace{4pt}\noindent\begin{minipage}{\columnwidth}
\begin{lstlisting}
Write the action exactly as one of the action templates, with every OBJ replaced by one of the listed objects. Only the listed objects can be used from here: to reach a room that is not listed, go to a listed room on the way, opening its door first if it is closed. If the last observation says that no known action matches, that action cannot work here, so do not repeat it. If the observation asks for a number to choose among actions, answer with that number alone.
\end{lstlisting}

\end{minipage}\par\vspace{4pt}

\textbf{WebShop.} The domain wording becomes ``WebShop online-shopping tasks of the product category'', and the skill is named \texttt{webshop-\{family\}}. The command list of the rewrite prompt is:
\par\vspace{4pt}\noindent\begin{minipage}{\columnwidth}
\begin{lstlisting}
search[{query}]; click[{product id}]; click[{option value}]; click[next >]; click[< prev]; click[back to search]; click[description]; click[features]; click[reviews]; click[buy now]
\end{lstlisting}

\end{minipage}\par\vspace{4pt}
Two sentences follow the command list:
\par\vspace{4pt}\noindent\begin{minipage}{\columnwidth}
\begin{lstlisting}
The skill serves every task of this category, and each task asks for a different product: do not write this task's product, attributes, option values or price into the skill; refer to them as what the instruction asks for.
How the pages work: a search opens a results page, where the way forward is to click a product id; "buy now" exists only on a product page. A product page lists its options (such as size, color or flavor) as clickable values: the options the instruction asks for must be clicked before "buy now". The purchase ends the task and is scored on product type, attributes, selected options and price.
\end{lstlisting}

\end{minipage}\par\vspace{4pt}

\textbf{LiveCodeBench.} There is no command list. The selector prompt opens with:
\par\vspace{4pt}\noindent\begin{minipage}{\columnwidth}
\begin{lstlisting}
An agent solves competitive programming problems in Python of type "{category title}": {category description}
It reads one reusable skill before every problem of this type. Keep the better of the two skills below, the one with which the agent's programs will pass the hidden tests more often on future problems of this type.
\end{lstlisting}

\end{minipage}\par\vspace{4pt}
The rewrite prompt uses the same domain wording, and the skill is named \texttt{livecode-\{category\}}. In place of a trajectory, it shows the problem statement, the program of the agent, and its results on the sample tests. The instruction to the generator becomes:
\par\vspace{4pt}\noindent\begin{minipage}{\columnwidth}
\begin{lstlisting}
Write the updated skill for future problems of this type, using what this attempt shows. The next problems are different problems of the same type, so the skill must not depend on this problem's statement.
\end{lstlisting}

\end{minipage}\par\vspace{4pt}

\subsection{Selector Training}
\label{sec:seltrain}
The selector backbone is Qwen3-4B. We train LoRA adapters with rank 32, scaling factor 64, and no dropout on all attention and feed-forward projections. We optimize with AdamW at a learning rate of $10^{-4}$ and without weight decay. The learning rate warms up linearly over the first 3\% of training and then follows a cosine decay. Gradients are clipped at norm 1.0. On ALFWorld, the private test set $\mathcal{H}_f$ of each family holds 100 tasks. The rejection threshold in Eq.~\eqref{eq:reject} is $\delta=0.1$, so the two skills of a pair differ by at least 10\% in success rate. Each batch holds 36 pairs, six from every family, and training makes one pass over the 24{,}000 training pairs. We use the final checkpoint. The validation skills serve for reporting only. They are not used to select a checkpoint or to tune hyperparameters. The backbone runs in BF16 with FlashAttention-2. The adapter weights, the label-token logits, the loss, and the optimizer states are kept in FP32.

\subsection{Inference}
\label{sec:inference}
At deployment, we merge the trained adapter into the backbone and serve the merged model. For each query, the server returns the log-probabilities of the two label tokens at the next position. Both values share the same normalization term. Comparing them is therefore equivalent to comparing the logits $z_A$ and $z_B$ in Eq.~\eqref{eq:accept}.

\subsection{Online Loop}
\label{sec:loop}
Algorithm~\ref{alg:online} gives the online selection loop of \S\ref{sec:online}. Every family starts from the empty skill. After each task, the generator rewrites the skill of the family, and the selector decides whether the rewrite replaces it.

\begin{algorithm}[!htb]
\caption{Online skill selection with \method{}}
\label{alg:online}
\begin{algorithmic}[1]
\Require frozen executor $E$, generator $G$, trained selector $C$, tasks $x_1,\dots,x_T$
\State $s_f\leftarrow\varnothing$ for every family $f$
\For{$t=1,\dots,T$}
  \State $f\leftarrow$ family of $x_t$; \ $(\tau,r)\leftarrow E(x_t,s_f)$
  \State $\tilde{s}\leftarrow G(s_f,\tau,r)$
  \If{$\tilde{s}$ passes the format check}
    \State place $s_f,\tilde{s}$ at A/B in random order
    \State $s_f\leftarrow$ the skill with the larger A/B logit under $C(f,\cdot,\cdot)$
  \EndIf
\EndFor
\end{algorithmic}
\end{algorithm}

\section{Further Experimental Results}
\label{sec:further}
\subsection{Results on WebShop}
\label{sec:webshop}
Table~\ref{tab:webshop} reports the online results on WebShop. The selector is trained with Qwen3-4B as the executor, and the Qwen3-8B columns change only the online executor. \method{} achieves the highest success rate and the highest score with both executors. With Qwen3-4B, it reaches 16.2\% success and a score of 55.9, against 14.7\% and 53.7 without skills. With Qwen3-8B, it reaches 18.5\% and 57.1, against 16.1\% and 54.9. The gains are smaller than on ALFWorld. A likely reason is that the Qwen3-4B generator lacks the capability to write high-quality skills for this environment, which limits what the selector can choose from.

\begin{table}[!htb]
\centering
\footnotesize
\setlength{\tabcolsep}{3pt}
\resizebox{\columnwidth}{!}{\begin{tabular}{lcccc}
\toprule
& \multicolumn{2}{c}{\textbf{Qwen3-4B}} & \multicolumn{2}{c}{\textbf{Qwen3-8B}} \\
\cmidrule(lr){2-3}\cmidrule(lr){4-5}
\textbf{Method} & Success & Score & Success & Score \\
\midrule
No-Skill & $14.7$ & $53.7$ & $16.1$ & $54.9$ \\
Accept-All & $15.8$ & $55.6$ & $18.1$ & $55.5$ \\
Random & $16.0$ & $55.0$ & $16.9$ & $54.3$ \\
LLM-Judge & $15.6$ & $55.3$ & $18.1$ & $55.3$ \\
LLM-Judge (Think) & $15.3$ & $55.1$ & $16.9$ & $54.8$ \\
LLM-Judge (Logit) & $15.5$ & $54.8$ & $17.5$ & $55.1$ \\
\rowcolor{oursrow}\method{} & $\mathbf{16.2}$ & $\mathbf{55.9}$ & $\mathbf{18.5}$ & $\mathbf{57.1}$ \\
\bottomrule
\end{tabular}
}
\caption{Online results on WebShop with two executors. Success is the success rate (\%) and Score is the mean purchase score. Bold marks the best value in each column.}
\label{tab:webshop}
\end{table}

\subsection{Combination with Other Methods}
\label{sec:combo}
\method{} only decides which of two skill documents to keep. It can therefore be attached to other skill evolution methods without further training. We attach it to SkillRise \citep{yao2026skillrise}, which trains one policy both to solve tasks and to curate the skill document. SkillRise keeps every curated document. With \method{}, the selector instead chooses between the old document and the curated one. The selector and its prompt are unchanged.

Table~\ref{tab:combo} reports the results on ALFWorld. \method{} improves our reproduction of SkillRise on both splits. Success rises from 86.0\% to 86.4\% on seen tasks and from 77.9\% to 80.1\% on unseen tasks. The selector thus also benefits a method whose executor it was not trained with.

\begin{table}[!htb]
\centering
\footnotesize
\setlength{\tabcolsep}{3pt}
\begin{tabular}{lcc}
\toprule
\textbf{Method} & \textbf{Seen} & \textbf{Unseen} \\
& (140) & (134) \\
\midrule
SkillRise (our reproduction) & $86.0$ & $77.9$ \\
\rowcolor{oursrow}\quad + \method{} & $\mathbf{86.4}$ & $\mathbf{80.1}$ \\
\bottomrule
\end{tabular}

\caption{\method{} attached to our reproduction of SkillRise on ALFWorld, reported as overall success rate (\%).}
\label{tab:combo}
\end{table}

\subsection{Efficiency}
\label{sec:efficiency}
Table~\ref{tab:efficiency} compares the cost of one selection. \method{} reads two logits from one forward pass and generates no tokens. It returns a decision in 24 ms. LLM-Judge generates about 16 tokens and needs 71 ms. LLM-Judge (Think) generates 1{,}144 tokens and needs 4{,}184 ms. \method{} is thus about 3 times faster than LLM-Judge and 174 times faster than LLM-Judge (Think).

\begin{table}[!htb]
\centering
\footnotesize
\setlength{\tabcolsep}{3pt}
\begin{tabular}{lcc}
\toprule
\textbf{Method} & \textbf{Tokens} & \textbf{Latency (ms)} \\
\midrule
LLM-Judge & 15.7 & 71 \\
LLM-Judge (Think) & 1{,}144 & 4{,}184 \\
LLM-Judge (Logit) & 0 & 26 \\
\rowcolor{oursrow}\method{} & 0 & 24 \\
\bottomrule
\end{tabular}

\caption{Cost of one selection during the online runs: generated tokens and latency.}
\label{tab:efficiency}
\end{table}

\section{Additional Related Work}
\label{sec:morerelated}
\paragraph{Skill learning and optimization.}
Beyond SkillRL and SkillRise, further methods combine policy learning with the construction or revision of a skill library \citep{huang2026skillselfplay,shi2026skill1,ge2026skillforge,zhang2026coevolvingskillgen,wang2025sage}. Other methods keep the executor fixed and revise the skill text through prompted optimization \citep{li2026skillhone,ni2026trace2skill,zhang2026coevoskills,wang2026skillgrad}. These methods focus on how a skill is produced or improved. However, our method does not change how skills are written. It only decides whether a rewrite replaces the current skill.

\paragraph{Skill construction and evaluation.}
LATM separates tool making from tool use \citep{cai2023latm}. A line of work constructs and reuses skills in different settings \citep{zheng2025skillweaver,yu2025polyskill,yang2026autoskill,jiang2026xskill,wang2026skillx,qiu2026autorefine,si2026ctx2skill}. Several surveys organize the area \citep{zhou2026skillsurvey,jiang2026sok,jiang2025adaptation}. SkillsBench and SkillGenBench benchmark the effect of skills and of skill generation pipelines \citep{li2026skillsbench,zhou2026skillgenbench}. An empirical study documents failures induced by skills \citep{dong2026skillsharmful}. These works construct or benchmark skills and show that a plausible skill can still hurt execution. However, our method uses the measured utility of skills as supervision for a selector that guards every online update.

\paragraph{Learned writers and skill selection.}
Skill-$\alpha$, SkillOS, and SKILLER learn or prompt the writer of the skill with feedback from a frozen executor \citep{shen2026skillalpha,ouyang2026skillos,dang2026skiller}. SkillMaster uses a counterfactual probe to decide which edits to retain during training \citep{yang2026skillmaster}. Other work learns when to invoke a skill or to which skill a query is routed \citep{guo2026skillapt,chen2026selskill,wang2026r3skill}. These methods shape how a skill is written or choose among the skills of a library. However, our method decides whether one rewrite replaces its predecessor.

\paragraph{Internalizing skills.}
SKILL0 withdraws contextual skills during reinforcement learning to internalize their behavior \citep{lu2026skill0}. Skill0.5 uses privileged distillation \citep{zhu2026skill05}. Context distillation trains a model to reproduce its prompted behavior without the prompt \citep{snell2022contextdistill,ye2026opcd}. These methods move guidance into model parameters. However, our method keeps the skill as text and leaves the executor unchanged.

\paragraph{Agent memory and context adaptation.}
Reflexion uses verbal feedback to improve subsequent attempts \citep{shinn2023reflexion}. ExpeL extracts insights from past trajectories \citep{zhao2023expel}. Agent Workflow Memory stores reusable workflows \citep{wang2024awm}. ReasoningBank organizes experience as reasoning memory \citep{ouyang2025reasoningbank}, and Buffer of Thoughts keeps reusable thought templates \citep{yang2024bot}. ACE treats context as an evolving resource \citep{zhang2025ace}. Training-Free GRPO and Metacognitive Reuse improve contextual guidance without updating model parameters \citep{cai2025trainingfreegrpo,didolkar2025metacognitive}. Evo-Memory evaluates evolving memory in sequential task streams \citep{wei2025evomemory}. These methods retrieve, accumulate, or consolidate general memory. However, our method keeps one skill document per task family and only decides between the current skill and one new skill.

\end{document}